\documentclass[11pt,letterpaper,UTF8]{article}
\usepackage{emnlp2017}
\usepackage{times}
\usepackage{CJKutf8}
\usepackage{bbm}
\usepackage{dsfont}
\usepackage{qtree}
\usepackage{multirow}
\usepackage{url}
\usepackage{latexsym}
\usepackage{graphicx}
\usepackage{tikz-qtree}
\usepackage{mathtools}

\usepackage{amsmath,amssymb,amsthm}

\newcommand{\R}{\ensuremath{\mathbb{R}}}

\newcommand{\ra}{\ensuremath{\rightarrow}}

\newcommand{\paren}[1]{\left(#1\right)}

\newcommand{\abs}[1]{\left|#1\right|}

\newcommand{\by}{\ensuremath{\times}}

\emnlpfinalcopy

\def\emnlppaperid{2}

\title{Entity Identification as Multitasking\Thanks{Part of the work was done while the author was at Bloomberg L.~P.}}

\author{
  {\bf Karl Stratos} \\
  Toyota Technological Institute at Chicago \\
  {\tt stratos@ttic.edu}
  }
\date{}

\begin{document}
\maketitle

\begin{abstract}

Standard approaches in entity identification hard-code boundary detection
and type prediction into labels and perform Viterbi.
This has two disadvantages:
1.~the runtime complexity grows quadratically in the number of types, and
2.~there is no natural segment-level representation.
In this paper, we propose a neural architecture that addresses these disadvantages.
We frame the problem as multitasking, separating boundary detection and type prediction but optimizing them jointly.
Despite its simplicity, this architecture performs competitively with fully structured models such as BiLSTM-CRFs
while scaling linearly in the number of types.
Furthermore, by construction, the model induces type-disambiguating embeddings of predicted mentions.

\end{abstract}

\section{Introduction}
\label{sec:intro}

A popular convention in segmentation tasks such as named-entity recognition (NER) and chunking is the so-called ``BIO''-label scheme.
It hard-codes boundary detection and type prediction into labels using the indicators ``B'' (Beginning), ``I'' (Inside), and ``O'' (Outside).
For instance, the sentence \texttt{Where is John Smith} is tagged as \texttt{Where/O is/O John/B-PER Smith/I-PER}.
In this way, we can treat the problem as sequence labeling and apply standard structured models such as CRFs.

But this approach has certain disadvantages.
First, the runtime complexity grows quadratically in the number of types (assuming exact decoding with first-order label dependency).
We emphasize that the asymptotic runtime remains quadratic even if we heuristically prune previous labels based on the BIO scheme.
This is not an issue when the number of types is small but quickly becomes problematic as the number grows.
Second, there is no segment-level prediction: every prediction happens at the word-level.
As a consequence, models do not induce representations corresponding to multi-word mentions,
which can be useful for downstream tasks such as named-entity disambiguation (NED).

In this paper, we propose a neural architecture that addresses these disadvantages.
Given a sentence, the model uses bidirectional LSTMs (BiLSTMs) to induce features and separately predicts:
\begin{enumerate}
\item Boundaries of mentions in the sentence.
\item Entity types of the boundaries.
\end{enumerate}
Crucially, during training, the errors of these two predictions are minimized jointly.

One might suspect that the separation could degrade performance;
neither prediction accounts for the correlation between entity types.
But we find that this is not the case due to joint optimization.
In fact, our model performs competitively with fully structured models such as BiLSTM-CRFs \cite{lample2016neural},
implying that the model is able to capture the entity correlation indirectly by multitasking.
On the other hand, the model scales linearly in the number of types
and induces segment-level embeddings of predicted mentions that are type-disambiguating by construction.

\section{Related Work}
\label{sec:related-work}

Our work is directly inspired by \newcite{lample2016neural}
who demonstrate that a simple neural architecture based on BiLSTMs
achieves state-of-the-art performance on NER with no external features.
They propose two models.
The first makes structured prediction of NER labels with a CRF loss (LSTM-CRF)
using the conventional BIO-label scheme.
The second, which performs slightly worse, uses a shift-reduce framework mirroring tansition-based dependency parsing \cite{yamada2003statistical}.
While the latter also scales linearly in the number of types and produces embeddings of predicted mentions, our approach is quite different.
We frame the problem as multitasking and do not need the stack/buffer data structure.
Semi-Markov models \cite{kong2015segmental,sarawagi2004semi} explicitly incorporate the segment structure
but are computationally intensive (quadratic in the sentence length).

Multitasking has been shown to be effective in numerous previous works \cite{collobert2011natural,yang2016multi,TACL885}.
This is especially true with neural networks which greatly simplify joint optimization across multiple objectives.
Most of these works consider multitasking across different problems.
In contrast, we decompose a single problem (NER) into two natural subtasks and perform them jointly.
Particularly relevant in this regard is the parsing model of \newcite{TACL885} which multitasks edge prediction and classification.

LSTMs \cite{hochreiter1997long}, and other variants of recurrent neural networks such as GRUs \cite{chung2014empirical},
have recently been wildly successful in various NLP tasks \cite{lample2016neural,TACL885,chung2014empirical}.
Since there are many detailed descriptions of LSTMs available, we omit a precise definition.
For our purposes, it is sufficient to treat an LSTM as a mapping
$\phi:\R^d \times \R^{d'} \ra \R^{d'}$ that takes an input vector $x$ and a state vector $h$
to output a new state vector $h' = \phi(x, h)$.

\section{Model}
\label{sec:model}

Let $\mathcal{C}$ denote the set of character types,
$\mathcal{W}$ the set of word types, and
$\mathcal{E}$ the set of entity types.
Let $\oplus$ denote the vector concatenation operation.
Our model first constructs a network over a sentence closely following \newcite{lample2016neural}; we describe it here for completeness.
The model parameters $\Theta$ associated with this base network are
\begin{itemize}
\item Character embedding $e_c \in \R^{25}$ for $c \in \mathcal{C}$
\item Character LSTMs $\phi^{\mathcal{C}}_f, \phi^{\mathcal{C}}_b: \R^{25} \times \R^{25} \ra \R^{25}$
\item Word embedding $e_w \in \R^{100}$ for $w \in \mathcal{W}$
\item Word LSTMs $\phi^{\mathcal{W}}_f, \phi^{\mathcal{W}}_b: \R^{150} \times \R^{100} \ra \R^{100}$
\end{itemize}
Let $w_1 \ldots w_n \in \mathcal{W}$ denote a word sequence where word $w_i$ has character $w_i(j) \in \mathcal{C}$ at position $j$.
First, the model computes a character-sensitive word representation $v_i \in \R^{150}$ as
\begin{align*}
f^{\mathcal{C}}_j &= \phi^{\mathcal{C}}_f\paren{e_{w_i(j)}, f^{\mathcal{C}}_{j-1}} &&\forall j = 1 \ldots \abs{w_i} \\
b^{\mathcal{C}}_j &= \phi^{\mathcal{C}}_b\paren{e_{w_i(j)}, b^{\mathcal{C}}_{j+1}} &&\forall j = \abs{w_i} \ldots 1\\
v_i &= f^{\mathcal{C}}_{\abs{w_i}} \oplus b^{\mathcal{C}}_1 \oplus e_{w_i} &&
\end{align*}
for each $i = 1 \ldots n$.\footnote{For simplicity, we assume some random initial
state vectors such as $f^{\mathcal{C}}_0$ and $b^{\mathcal{C}}_{\abs{w_i}+1}$ when we describe LSTMs.}
Next, the model computes
\begin{align*}
f^{\mathcal{W}}_i &= \phi^{\mathcal{W}}_f\paren{v_i, f^{\mathcal{W}}_{i-1}}  &&\forall i = 1 \ldots n \\
b^{\mathcal{W}}_i &= \phi^{\mathcal{W}}_b\paren{v_i, b^{\mathcal{W}}_{i+1}} &&\forall i = n \ldots 1
\end{align*}
and induces a character- and context-sensitive word representation $h_i \in \R^{200}$ as
\begin{align}
h_i &= f^{\mathcal{W}}_i \oplus b^{\mathcal{W}}_i \label{eq:h}
\end{align}
for each $i = 1 \ldots n$. These vectors are used to define the boundary detection loss and the type classification loss described below.

\paragraph{Boundary detection loss}
We frame boundary detection as predicting BIO tags without types.
A natural approach is to optimize the conditional probability of the correct tags
$y_1 \ldots y_n \in \{\texttt{B}, \texttt{I}, \texttt{O}\}$:
\begin{align}
  p(y_1 \ldots& y_n |  h_1 \ldots h_n) \notag\\
  &\propto \exp\paren{\sum_{i=1}^n T_{y_{i-1}, y_i} \times g_{y_i}(h_i)} \label{eq:crf}
\end{align}
where $g: \R^{200} \ra \R^3$ is a function that adjusts the length of the LSTM output to the number of targets.
We use a feedforward network $g(h) = W^2 \mbox{relu}(W^1 h + b^1) + b^2$.
We write $\Theta_1$ to refer to $T \in \R^{3 \by 3}$ and the parameters in $g$.
The boundary detection loss is given by the negative log likelihood:
\begin{align*}
L_1\paren{\Theta, \Theta_1} = - \sum_l \log p\paren{y^{(l)} | h^{(l)}}
\end{align*}
where $l$ iterates over tagged sentences in the data.

The global normalizer for \eqref{eq:crf} can be computed using dynamic programming; see \newcite{collobert2011natural}.
Note that the runtime complexity of boundary detection is constant despite dynamic programming since the number of tags is fixed (three).

\paragraph{Type classification loss}
Given a mention boundary $1\leq s \leq t \leq n$, we predict its type using \eqref{eq:h} as follows.
We introduce an additional pair of LSTMs $\phi^{\mathcal{E}}_f, \phi^{\mathcal{E}}_b: \R^{200} \times \R^{200} \ra \R^{200}$ and compute
a corresponding mention representation $\mu \in \R^{\abs{\mathcal{E}}}$ as
\begin{align}
f^{\mathcal{E}}_j &= \phi^{\mathcal{E}}_f\paren{h_j, f^{\mathcal{E}}_{j-1}} &&\forall j = s \ldots t  \notag \\
b^{\mathcal{E}}_j &= \phi^{\mathcal{E}}_b\paren{h_j, b^{\mathcal{E}}_{j+1}} &&\forall j = t \ldots s  \notag \\
\mu &= q\paren{f^{\mathcal{E}}_t \oplus b^{\mathcal{E}}_s} && \label{eq:mention}
\end{align}
where $q: \R^{400} \ra \R^{\abs{\mathcal{E}}}$ is again a feedforward network that adjusts the vector length to $\abs{\mathcal{E}}$.\footnote{Clearly,
one can consider different networks over the boundary, for instance simple bag-of-words or convolutional neural networks. We leave the exploration as future work.}
We write $\Theta_2$ to refer to the parameters in $\phi^{\mathcal{E}}_f, \phi^{\mathcal{E}}_b, q$.
Now we can optimize the conditional probability of the correct type $\tau$:
\begin{align}
p(\tau | h_s \ldots h_t) \propto \exp\paren{\mu_\tau} \label{eq:class}
\end{align}
The type classification loss is given by the negative log likelihood:
\begin{align*}
L_2\paren{\Theta, \Theta_2} = - \sum_l \log p\paren{\tau^{(l)} | h^{(l)}_s \ldots h^{(l)}_t}
\end{align*}
where $l$ iterates over typed mentions in the data.

\paragraph{Joint loss}
The final training objective is to minimize the sum of the boundary detection loss and the type classification loss:
\begin{align}
L(\Theta, \Theta_1, \Theta_2) = L_1\paren{\Theta, \Theta_1} + L_2\paren{\Theta, \Theta_2} \label{eq:joint}
\end{align}
In stochastic gradient descent (SGD), this amounts to computing the tagging loss $l_1$
and the classification loss $l_2$ (summed over all mentions) at each annotated sentence, and then taking a gradient step on $l_1 + l_2$.
Observe that the base network $\Theta$ is optimized to handle both tasks.
During training, we use gold boundaries and types to optimize $L_2\paren{\Theta, \Theta_2}$.
At test time, we predict boundaries from the tagging layer \eqref{eq:crf}
and classify them using the classification layer \eqref{eq:class}.

\section{Experiments}
\label{sec:experiments}

\paragraph{Data}
We use two NER datasets: CoNLL 2003 which has four entity types \texttt{PER}, \texttt{LOC}, \texttt{ORG} and \texttt{MISC} \cite{tjong2003introduction},
and the newswire portion of OntoNotes Release 5.0 which has 18 entity types \cite{weischedel2013ontonotes}.

\paragraph{Implementation and baseline} We denote our model Mention2Vec and implement it
using the DyNet library.\footnote{\url{https://github.com/karlstratos/mention2vec}}
We use the same pre-trained word embeddings in \newcite{lample2016neural}.
We use the Adam optimizer \cite{kingma2014adam} and apply dropout at all LSTM layers \cite{hinton2012improving}.
We perform minimal tuning over development data. Specifically, we perform a $5 \by 5$ grid search over learning rates $0.0001 \ldots 0.0005$
and dropout rates $0.1 \ldots 0.5$ and choose the configuration that gives the best performance on the dev set.

\begin{table}[t!]
\begin{center}
{
\begin{tabular}{|l|l|c|}
\hline
CoNLL 2003 (4 types)               &   F1      &  \# words/sec\\
\hline
BiLSTM-CRF                         &  90.22    &   3889 \\
Mention2Vec                        &  90.90    &   4825 \\
\hline
\hline
OntoNotes (18 types)               &   F1      &  \# words/sec\\
\hline
BiLSTM-CRF                         &  90.77    &   495  \\
Mention2Vec                        &  89.37    &   4949 \\
\hline
\end{tabular}
\caption{Test F1 scores on CoNLL 2003 and OntoNotes newswire portion.}
\label{tab:comp1}
}
\end{center}
\vspace{-3mm}
\end{table}

\begin{table}[t!]
\begin{center}
{
\begin{tabular}{|l|l|}
\hline
Model                              &   F1   \\
\hline
\newcite{mccallum2003early}        &  84.04 \\
\newcite{collobert2011natural}     &  89.59 \\
\newcite{lample2016neural}--Greedy &  89.15 \\
\newcite{lample2016neural}--Stack  &  90.33 \\
\newcite{lample2016neural}--CRF    &  90.94 \\
\hline
Mention2Vec                        &  90.90 \\
\hline
\end{tabular}
\caption{Test F1 scores on CoNLL 2003.}
\label{tab:comp2}
}
\end{center}
\vspace{-5mm}
\end{table}

\begin{table*}[t!]
\begin{center}
{\footnotesize
\begin{tabular}{|l|l|}
\hline
\texttt{PER} & In another letter dated January 1865, a well-to-do Washington matron wrote to \textbf{Lincoln} to plead for $\ldots$ \\
\hline
& Chang and \textbf{Washington} were the only men's seeds in action on a day that saw two seeded women's $\ldots$ \\
& ``Just one of those things, I was just trying to make contact,'' said \textbf{Bragg}. \\
& \textbf{Washington}'s win was not comfortable, either. \\
\hline
\texttt{LOC} & Lauck, from \textbf{Lincoln}, Nebraska, yelled a tirade of abuse at the court after his conviction for inciting $\ldots$ \\
\hline
&$\ldots$ warring factions, with the PUK aming to break through to KDP's headquarters in \textbf{Saladhuddin}. \\
&$\ldots$ is not expected to travel to the \textbf{West Bank} before Monday,'' Nabil Abu Rdainah told Reuters. \\
&$\ldots$ off a bus near his family home in the village of \textbf{Donje Ljupce} in the municipality of Podujevo. \\
\hline
\texttt{ORG} & English division three - Swansea v \textbf{Lincoln}. \\
\hline
&SOCCER - OUT-OF-SORTS \textbf{NEWCASTLE} CRASH 2 1 AT HOME. \\
&Moura, who appeared to have elbowed Cyprien in the final minutes of the 3 0 win by \textbf{Neuchatel}, was $\ldots$ \\
&In Sofia: Leviski Sofia (Bulgaria) 1 \textbf{Olimpija} (Slovenia) 0 \\
\hline
\texttt{WORK\_OF\_ART} &  $\ldots$ Bond novels, and ``\textbf{Treasure Island},'' produced by Charlton Heston who also stars in the movie. \\
\hline
&$\ldots$ probably started in 1962 with the publication of Rachel Carson's book ``\textbf{Silent Spring}.'' \\
&$\ldots$ Victoria Petrovich) spout philosophic bon mots with the self-concious rat-a-tat pacing of ``\textbf{Laugh In}.'' \\
&Dennis Farney's Oct. 13 page - one article ``\textbf{River of Despair},'' about the poverty along the $\ldots$ \\
\hline
\texttt{GPE} & $\ldots$ from a naval station at \textbf{Treasure Island} near the Bay Bridge to San Francisco to help fight fires. \\
\hline
&$\ldots$ lived in an expensive home on \textbf{Lido Isle}, an island in Newport's harbor, according to investigators. \\
&$\ldots$ Doris Moreno, 37, of \textbf{Bell Gardens}; and Ana L. Azucena, 27, of Huntington Park. \\
&One group of middle-aged manufacturing men from the company's \textbf{Zama} plant outside Tokyo was $\ldots$ \\
\hline
\texttt{ORG} & $\ldots$ initiative will spur members of \textbf{the General Agreement on Tariffs and Trade} to reach $\ldots$ \\
\hline
& $\ldots$ question of Taiwan's membership in \textbf{the General Agreement on Tariffs and Trade} should $\ldots$ \\
&''He doesn't know himself,'' Kathy Stanwick of \textbf{the Abortion Rights League} says of $\ldots$ \\
&$\ldots$ administrative costs, management and research, \textbf{the Office of Technology Assessment} just reported.\\
\hline

\end{tabular}
\caption{Nearest neighbors of detected mentions in CoNLL 2003 and OntoNotes using \eqref{eq:mention}.}
\label{tab:neighbors}
}
\end{center}
\vspace{-3mm}
\end{table*}

We also re-implement the BiLSTM-CRF model of \newcite{lample2016neural};
this is equivalent to optimizing just $L_1(\Theta, \Theta_1)$ but using typed BIO tags.
\newcite{lample2016neural} use different details
in optimization (SGD with gradient clipping), data preprocessing (replacing every digit with a zero),
and the dropout scheme (droptout at BiLSTM input \eqref{eq:h}).
As a result, our re-implementation is not directly comparable and obtains different (slightly lower) results.
But we emphasize that the main goal of this paper is to demonstrate the utility the proposed approach
rather than obtaining a new state-of-the-art result on NER.

\subsection{NER Performance}

Table~\ref{tab:comp1} compares the NER performance and decoding speed between BiLSTM-CRF and Mention2Vec.
The F1 scores are obtained on test data.
The speed is measured by the average number of words decoded per second.

On CoNLL 2003 in which the number of types is small, our model achieves 90.50 compared to 90.22 of BiLSTM-CRF with minor speed improvement.
This shows that despite the separation between boundary detection and type classification, we can achieve good performance through joint optimization.
On OntoNotes in which the number of types is much larger, our model still performs well with an F1 score of 89.37 but is behind
BiLSTM-CRF which achieves 90.77. We suspect that this is due to strong correlation between mention types that fully structured models
can exploit more effectively.
However, our model is also an order of magnitude faster: 4949 compared to 495 words/second.

Finally, Table~\ref{tab:comp2} compares our model with other works in the literature on CoNLL 2003.
\newcite{mccallum2003early} use CRFs with manually crafted features;
\newcite{collobert2011natural} use convolutional neural networks;
\newcite{lample2016neural} use BiLSTMs in a greedy tagger (Greedy),
a stack-based model (Stack), and a global tagger using a CRF output layer (CRF).
Mention2Vec performs competitively.

\subsection{Mention Embeddings}

Table~\ref{tab:neighbors} shows nearest neighbors of detected mentions using the mention representations $\mu$ in \eqref{eq:mention}.
Since $\mu_\tau$ represents the score of type $\tau$, the mention embeddings are clustered by entity types \textit{by construction}.
The model induces completely different representations even when the mention has the same lexical form.
For instance, based on its context \texttt{Lincoln} receives a person, location, or organization representation;
\texttt{Treasure Island} receives a book or location representation. The model also learns representations for long multi-word expressions such as
\texttt{the General Agreement on Tariffs and Trade}.

\section{Conclusion}
\label{sec:conclusion}

We have presented a neural architecture for entity identification that multitasks boundary detection and type classification.
Joint optimization enables the base BiLSTM network to capture the correlation between entities indirectly via multitasking.
As a result, the model is competitive with fully structured models such as BiLSTM-CRFs on CoNLL 2003
while being more scalable and also inducing context-sensitive mention embeddings clustered by entity types.
There are many interesting future directions, such as applying this framework to NED in which type classification is much more fine-grained
and finding a better method for optimizing the multitasking objective
(e.g., instead of using gold boundaries for training, dynamically use predicted boundaries in a reinforcement learning framework).

\section*{Acknowledgments}
The author would like to thank Linpeng Kong for his consistent help with DyNet
and Miguel Ballesteros for pre-trained word embeddings.

\bibliography{mention2vec}

\begin{thebibliography}{}
\expandafter\ifx\csname natexlab\endcsname\relax\def\natexlab#1{#1}\fi

\bibitem[{Chung et~al.(2014)Chung, Gulcehre, Cho, and
  Bengio}]{chung2014empirical}
Junyoung Chung, Caglar Gulcehre, KyungHyun Cho, and Yoshua Bengio. 2014.
\newblock Empirical evaluation of gated recurrent neural networks on sequence
  modeling.
\newblock In {\em NIPS Deep Learning Workshop\/}.

\bibitem[{Collobert et~al.(2011)Collobert, Weston, Bottou, Karlen, Kavukcuoglu,
  and Kuksa}]{collobert2011natural}
Ronan Collobert, Jason Weston, L{\'e}on Bottou, Michael Karlen, Koray
  Kavukcuoglu, and Pavel Kuksa. 2011.
\newblock Natural language processing (almost) from scratch.
\newblock {\em The Journal of Machine Learning Research\/} 12:2493--2537.

\bibitem[{Hinton et~al.(2012)Hinton, Srivastava, Krizhevsky, Sutskever, and
  Salakhutdinov}]{hinton2012improving}
Geoffrey~E Hinton, Nitish Srivastava, Alex Krizhevsky, Ilya Sutskever, and
  Ruslan~R Salakhutdinov. 2012.
\newblock Improving neural networks by preventing co-adaptation of feature
  detectors.
\newblock {\em arXiv preprint arXiv:1207.0580\/} .

\bibitem[{Hochreiter and Schmidhuber(1997)}]{hochreiter1997long}
Sepp Hochreiter and J{\"u}rgen Schmidhuber. 1997.
\newblock Long short-term memory.
\newblock {\em Neural computation\/} 9(8):1735--1780.

\bibitem[{Kingma and Ba(2014)}]{kingma2014adam}
Diederik Kingma and Jimmy Ba. 2014.
\newblock Adam: A method for stochastic optimization.
\newblock {\em arXiv preprint arXiv:1412.6980\/} .

\bibitem[{Kiperwasser and Goldberg(2016)}]{TACL885}
Eliyahu Kiperwasser and Yoav Goldberg. 2016.
\newblock Simple and accurate dependency parsing using bidirectional lstm
  feature representations.
\newblock {\em Transactions of the Association for Computational Linguistics\/}
  4:313--327.

\bibitem[{Kong et~al.(2015)Kong, Dyer, and Smith}]{kong2015segmental}
Lingpeng Kong, Chris Dyer, and Noah~A Smith. 2015.
\newblock Segmental recurrent neural networks.
\newblock {\em arXiv preprint arXiv:1511.06018\/} .

\bibitem[{Lample et~al.(2016)Lample, Ballesteros, Subramanian, Kawakami, and
  Dyer}]{lample2016neural}
Guillaume Lample, Miguel Ballesteros, Sandeep Subramanian, Kazuya Kawakami, and
  Chris Dyer. 2016.
\newblock Neural architectures for named entity recognition.
\newblock In {\em Proceedings of NAACL\/}.

\bibitem[{McCallum and Li(2003)}]{mccallum2003early}
Andrew McCallum and Wei Li. 2003.
\newblock Early results for named entity recognition with conditional random
  fields, feature induction and web-enhanced lexicons.
\newblock In {\em Proceedings of the seventh conference on Natural language
  learning at HLT-NAACL 2003-Volume 4\/}. Association for Computational
  Linguistics, pages 188--191.

\bibitem[{Plank(2016)}]{plank1}
Barbara Plank. 2016.
\newblock Keystroke dynamics as signal for shallow syntactic parsing.
\newblock In {\em Proceedings of COLING\/}.

\bibitem[{Plank et~al.(2016)Plank, S{\o}gaard, and Goldberg}]{plank2}
Barbara Plank, Anders S{\o}gaard, and Yoav Goldberg. 2016.
\newblock Multilingual part-of-speech tagging with bidirectional long
  short-term memory models and auxiliary loss.
\newblock In {\em Proceedings of ACL\/}.

\bibitem[{Sarawagi et~al.(2004)Sarawagi, Cohen et~al.}]{sarawagi2004semi}
Sunita Sarawagi, William~W Cohen, et~al. 2004.
\newblock Semi-markov conditional random fields for information extraction.
\newblock In {\em NIPs\/}. volume~17, pages 1185--1192.

\bibitem[{Tjong Kim~Sang and De~Meulder(2003)}]{tjong2003introduction}
Erik~F Tjong Kim~Sang and Fien De~Meulder. 2003.
\newblock Introduction to the conll-2003 shared task: Language-independent
  named entity recognition.
\newblock In {\em Proceedings of the seventh conference on Natural language
  learning at HLT-NAACL 2003-Volume 4\/}. Association for Computational
  Linguistics, pages 142--147.

\bibitem[{Weischedel et~al.(2013)Weischedel, Palmer, Marcus, Hovy, Pradhan,
  Ramshaw, Xue, Taylor, Kaufman, Franchini et~al.}]{weischedel2013ontonotes}
Ralph Weischedel, Martha Palmer, Mitchell Marcus, Eduard Hovy, Sameer Pradhan,
  Lance Ramshaw, Nianwen Xue, Ann Taylor, Jeff Kaufman, Michelle Franchini,
  et~al. 2013.
\newblock Ontonotes release 5.0 ldc2013t19.
\newblock {\em Linguistic Data Consortium, Philadelphia, PA\/} .

\bibitem[{Yamada and Matsumoto(2003)}]{yamada2003statistical}
Hiroyasu Yamada and Yuji Matsumoto. 2003.
\newblock Statistical dependency analysis with support vector machines.
\newblock In {\em Proceedings of IWPT\/}. volume~3, pages 195--206.

\bibitem[{Yang et~al.(2016)Yang, Salakhutdinov, and Cohen}]{yang2016multi}
Zhilin Yang, Ruslan Salakhutdinov, and William Cohen. 2016.
\newblock Multi-task cross-lingual sequence tagging from scratch.
\newblock {\em arXiv preprint arXiv:1603.06270\/} .

\end{thebibliography}
\bibliographystyle{emnlp_natbib}

\end{document}